\pdfoutput=1
\documentclass[11pt]{article}

\usepackage[preprint]{acl}

\usepackage{times}
\usepackage{latexsym}
\usepackage{algorithm}
\usepackage[noEnd=true,rightComments=false]{algpseudocodex}

\usepackage[T1]{fontenc}
\usepackage{booktabs}
\usepackage[utf8]{inputenc}

\usepackage{microtype}

\usepackage{inconsolata}

\usepackage{graphicx}
\usepackage{subcaption}
\usepackage{multirow}
\usepackage{listings}
\usepackage{amssymb, amsmath}
\usepackage{xcolor}

\title{Code-Optimise: Self-Generated Preference Data \\for Correctness and Efficiency}

\author{
    Leonidas Gee\textsuperscript{\normalfont1}\thanks{\ Research internship at Huawei Noah's Ark Lab, London.}, Milan Gritta\textsuperscript{\normalfont2}, Gerasimos Lampouras\textsuperscript{\normalfont2}, and Ignacio Iacobacci\textsuperscript{\normalfont2} \\
    \textsuperscript{1}University of Sussex, Brighton, UK \\ \textsuperscript{2}Huawei Noah’s Ark Lab, London, UK \\
    \texttt{jg717@sussex.ac.uk} \\ \texttt{\{milan.gritta, gerasimos.lampouras, ignacio.iacobacci\}@huawei.com}
}

\begin{document}
\maketitle

%%% Abstract %%%

\begin{abstract}

Code Language Models have been trained to generate accurate solutions, typically with no regard for runtime. On the other hand, previous works that explored execution optimisation have observed corresponding drops in functional correctness. To that end, we introduce \textbf{Code-Optimise}, a framework that incorporates both correctness (passed, failed) and runtime (quick, slow) as learning signals via \textit{self-generated preference data}. Our framework is both lightweight and robust as it dynamically selects solutions to reduce overfitting while avoiding a reliance on larger models for learning signals. Code-Optimise achieves significant improvements in $pass@k$ while decreasing the competitive baseline runtimes by an additional 6\% for in-domain data and up to 3\% for out-of-domain data. As a by-product, the average length of the generated solutions is reduced by up to 48\% on MBPP and 23\% on HumanEval, resulting in faster and cheaper inference. The generated data and codebase is open-sourced at \url{https://github.com/huawei-noah/HEBO/tree/Code_Optimise}.

\end{abstract}

\begin{figure}[t!]
    \centering
        \includegraphics[scale=0.68]{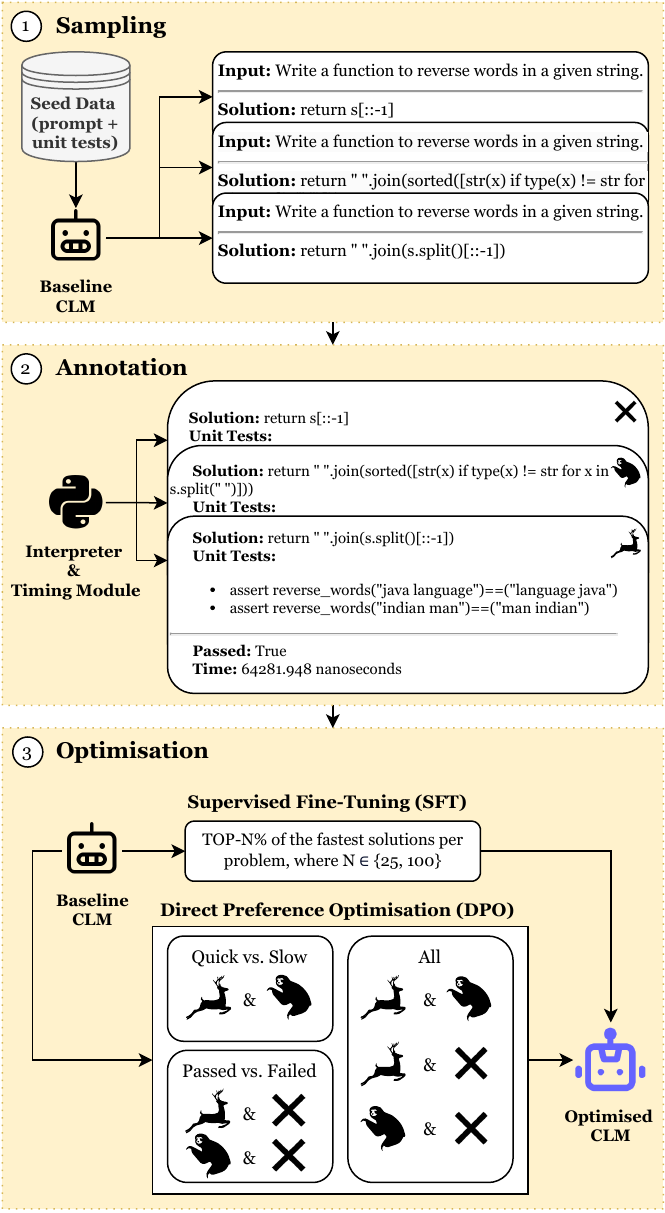}

        \caption{Overview of Code-Optimise. (1) Diverse solutions are sampled per problem. (2) A code interpreter annotates the solutions by functional correctness and runtime. (3) CLM is optimised using SFT or DPO.}
    \label{fig:schema}
\end{figure}

\begin{table*}[t]
    \centering
        \begin{tabular}{l|c|ccc|cccc}
            \toprule

            \multicolumn{1}{c|}{\multirow{2}{*}{\textbf{Model}}}    & \multirow{2}{*}{\textbf{Split}}   & \multicolumn{3}{c|}{\textbf{Problem}} & \multicolumn{4}{c}{\textbf{Solution}}    \\

            &  & \textbf{Total}    & \textbf{Filtered}    & \textbf{Ratio}    & \textbf{Total}    & \textbf{Filtered}    & \textbf{Ratio}    & \textbf{CoV}   \\

            \midrule

            \multirow{2}{*}{StarCoder-1B}   & Train         & 384       & 183       & 47.66      & 38400        & 15472     & 40.29     & 0.011   \\
                                            & Validation    & 90        & 40        & 44.44      & 9000         & 3533      & 39.26     & 0.010   \\

            \multirow{2}{*}{StarCoder-3B}   & Train         & 384       & 211       & 54.95      & 38400        & 17575     & 45.77     & 0.007   \\
                                            & Validation    & 90        & 45        & 50.00      & 9000         & 3926      & 43.62     & 0.014   \\

            \multirow{2}{*}{CodeLlama-7B}   & Train         & 384       & 250       & 65.10      & 38400        & 21350     & 55.60     & 0.007   \\
                                            & Validation    & 90        & 55        & 61.11      & 9000         & 4962      & 55.13     & 0.008   \\

            \multirow{2}{*}{CodeLlama-13B}  & Train         & 384       & 261       & 67.97      & 38400        & 22182     & 57.77     & 0.007   \\
                                            & Validation    & 90        & 56        & 62.22      & 9000         & 5108      & 56.76     & 0.007   \\

            \bottomrule
        \end{tabular}

    \caption{Statistics of our self-generated preference data. 1) A \textbf{Model} generates 100 solutions per problem out of the \textbf{Total} problems in each \textbf{Split}. 2) Functional correctness and runtime are annotated. 3) Problems are filtered to retain those with at least 2 passing and 1 failing solution (\textbf{Filtered}). A low coefficient of variation (\textbf{CoV} $\le$ 0.1) across 5 runs indicates that runtime measurements are stable. \textbf{Ratio} is the percentage of \( \frac{\textbf{Filtered}}{\textbf{Total}} \) retained code solutions.}
    \label{tab:synthetic}
\end{table*}

%%% Introduction %%%

\section{Introduction}

Code Language Models (CLMs) trained on large code repositories such as The Stack \cite{kocetkov2022stack, lozhkov2024starcoder} gradually increase their understanding of code semantics. CLMs are thus able to generate functionally correct and reasonably efficient solutions to programming problems \cite{austin2021program, chen2021evaluating}, among many other code-related skills \cite{li2023starcoder}. \citet{shypula2023learning} have shown that CLMs can optimise slow-running code to achieve large runtime gains but at a substantial cost to correctness (down by up to $\sim$30\%). Subsequent research has focused mostly on improving code correctness. From the data perspective, a common way of improving functional correctness is via distilled supervised fine-tuning \cite{tunstall2023zephyr, xu2023wizardlm, luo2023wizardcoder, wei2023magicoder} on training data generated by large models such as GPT-4 \cite{achiam2023gpt}. However, in many cases, due to legal, financial and/or privacy constraints, it is not feasible to rely on proprietary data. Furthermore, we seek to overcome the limitations of supervised fine-tuning (SFT), which only optimises for `positive' examples and possesses no means of \textit{reducing the likelihood} that undesirable (e.g. incorrect or slow) code is generated. Although such issues may be addressed via Reinforcement Learning (RL;  \citet{le2022coderl, wang2022compilable, gorinski2023automatic}), they often come with added complexity and instability. Therefore, we opt for Direct Preference Optimisation \cite{rafailov2024direct} as our preferred fine-tuning method due to its simplicity and widespread adoption. 

We propose \textbf{Code-Optimise}, a lightweight framework that trains CLMs with our \textit{self-generated preference data} for correctness (passed / failed) and efficiency (quick / slow), shown in Figure \ref{fig:schema}. Starting from a small collection of problems and unit tests, Code-Optimise bootstraps the pre-trained CLM to generate the required learning signals, thereby exposing the model to automatically annotated on-policy data. Additional robustness is also provided by dynamically selecting solutions during training to reduce overfitting. Our method consists of three steps: 1) \textit{Sampling}; generate \textit{N} solutions for each problem description, 2) \textit{Annotation}; automatically label each solution for correctness and runtime, 3) \textit{Optimisation}; fine-tune the CLM on self-generated preference data using several lightweight configurations. Our main contributions are:

\begin{itemize}

  \item We create and publish a novel code preference dataset (and recipe to extend it) that enables multi-objective optimisation (code correctness and runtime efficiency) of CLMs.

  \item We present experimental analysis to support our approach and observe that functional correctness is significantly improved, particularly for smaller CLMs and lower $k$ in $pass@k$. The scores are further enhanced with our Dynamic Solution Selection (DSS). 

  \item We demonstrate that runtimes are reduced by up to 6\% for MBPP and 3\% for HumanEval over competitive baseline CLMs. Finally, the length of generated solutions is reduced by up to 23\% for HumanEval and 48\% for MBPP, thereby decreasing inference costs.

\end{itemize}

\noindent To the best of our knowledge, our work is the first to show improvements in both correctness \textit{and} efficiency for the task of code generation.

%%% Methodology %%%

\section{Code-Optimise}\label{sec:methodology}

In Figure \ref{fig:schema}, we introduce \textbf{Code-Optimise}, a lightweight optimisation method for CLMs aimed at improving functional correctness and/or reducing the runtime of code. We assume that the pre-trained model has knowledge of accurate and efficient solutions but requires guidance to determine the best solution. This is also one of the main motivations of preference optimisation algorithms. Unlike training with distilled signals from larger models, new information is not learned. These approaches may be orthogonal. However, such an exploration is beyond the scope of this work.

\subsection{Sampling}\label{sec:sampling}

We assume access to \(D_{seed}=\{x_i, y_i, ut_i\}_{i=1}^{N}\), a dataset of problem descriptions $x_i$ and the corresponding unit tests $ut_i$ that can be used for sampling and testing new solutions from the CLM, denoted $CLM_{base}$ henceforth. Since fine-tuning the model on the limited solutions $y_i$ would lead to overfitting, we leverage its extensive pre-training to generate a \textit{multitude} of diverse solutions to obtain additional training data. We sample 100 solutions from $CLM_{base}$ for each problem description with multinomial sampling due to its lower computational cost. A temperature of $t=0.6$ is applied to achieve a balance between functional correctness \textit{and} diversity, resulting in non-uniform runtimes.

\begin{algorithm}[t]
\caption{Timing module algorithm.}
\label{alg:timing}
    \begin{algorithmic}[1]
        \For{$s \in solutions$}
          \State $CoV \gets \infty$
          \Repeat \ \Comment{up to 1K times}
            \State $times \gets [\ ]$ \Comment{initialise empty list}
            \For{$1, \dots, 50$}
              \State $runtime, passed \gets \Call{exec}{s}$
              \State $times.append(runtime)$
            \EndFor
              \State $\mu, \sigma \gets \Call{mean}{times}, \Call{std}{times}$
              \State $CoV \gets \sigma / \mu$
          \Until{$CoV \leq 0.1$}
          \If{$CoV > 0.1$} 
            \State \Comment{stable runtime was not obtained}
            \State $passed \gets False$
          \EndIf
        \EndFor
    \end{algorithmic}
\end{algorithm}

\subsection{Annotation}

The solutions are automatically evaluated for functional correctness and runtime. While the former can be achieved by simply executing a solution with its corresponding unit tests, the latter requires additional steps for obtaining stable runtime measurements, see Algorithm \ref{alg:timing}. Each solution $s$ is executed 50 times to determine its functional correctness (passed/failed) and $runtime$ in nanoseconds. We obtain $\mu$ and $\sigma$, then calculate the coefficient of variation $CoV$. A measurement is deemed stable and accepted if $CoV \le 0.1$ (usually much lower). Otherwise, we repeat the loop up to 1K times. In the \textit{unlikely} scenario that a stable $runtime$ could not be obtained, we set $passed=False$ (mark solution as failed). In order to further increase the reliability of $runtime$ measurements, we execute Algorithm \ref{alg:timing} five times (in a \textit{separate process}) and average the results. Lastly, we remove problems {$x_i, y_i, ut_i$} which do not have at least \textit{two} passing and \textit{one} failed solution to ensure that optimisation can be enhanced with our Dynamic Solution Selection (\ref{dynamic_selection}) during training. The statistics of the final dataset $D_{train}$ are shown in Table \ref{tab:sampling}.

\subsection{Optimisation}

In this step, the model is fine-tuned on $D_{train}$ to bias $CLM_{base}$ towards generating more functionally correct and runtime-efficient solutions. Although several methods for preference data optimisation exist \citep{yuan2023rrhf, zhao2023slic, liu2023statistical, azar2024general, ethayarajh2024kto, hong2024orpo}, we opt for DPO due to its simplicity and wide adoption. We also benchmark SFT due to its widespread use in prior work.

\paragraph{Supervised Fine-Tuning}

We train $CLM_{base}$ on $D_{train}$ using TOP-$N$\% of the fastest solutions where $N \in \{ 25, 100 \}$, which means that the diversity of runtimes grows as $N$ increases. Henceforth, models optimised with the top 25\% of fastest solutions are denoted as $SFT_{25}$ and CLMs trained with all (including the \textit{slowest}) solutions as $SFT_{100}$.  
% deliberately no space here for better formatting
\begin{equation}
\label{eq:sft}
    \mathcal{L}_{\mathrm{SFT}} \left( \pi_{\theta} \right) = -\mathbb{E}_{\left( x, y \right) \sim D} \left[ \log \pi_{\theta} \left(y \mid x \right) \right]
\end{equation}

\paragraph{Direct Preference Optimisation}

Aiming to avoid the complexity and instability of reinforcement learning, DPO \citep{rafailov2024direct} aligns  models to preference data with a simple classification loss, shown in Equation \ref{eq:dpo}.
\begin{multline}
\label{eq:dpo}
  \mathcal{L}_{\mathrm{DPO}} \left( \pi_\theta; \pi_{\mathrm{ref}} \right) = -\mathbb{E}_{\left( x, y_{w}, y_{l} \right) \sim \mathcal{D}} \\
    \log \sigma \left( \beta \log \frac{ \pi_{\theta} \left( y_{w} \mid x \right) }{ \pi_{\mathrm{ref}} \left( y_{w} \mid x \right) } - \beta \log \frac{ \pi_{\theta} \left( y_{l} \mid x \right) }{ \pi_{\mathrm{ref}} \left( y_{l} \mid x \right) } \right)
\end{multline}

\noindent We investigate the effectiveness of the following configurations of code preference pairs:

\begin{itemize}
  \item \textbf{Quick versus Slow:} Choose \textit{quick \& slow} solutions according to the annotated runtime. We denote such models as $DPO_{QvS}$.
  % \footnote{NB: Only correct code can be assigned a runtime. Thus, functional correctness and code efficiency \textit{cannot be separated} as learning signals. The $DPO_{QvS}$ configuration \textit{implicitly} optimises $CLM_{base}$ for correctness as well.} 

  \item \textbf{Passed versus Failed:} Choose \textit{passed \& failed} pairs according to the annotated functional correctness, denoted as $DPO_{PvF}$.

  \item \textbf{All:} Choose all preference pairs from the \textit{Quick vs. Slow} and \textit{Passed vs. Failed} configurations. We denote such models as $DPO_{All}$.
\end{itemize}

\begin{figure*}[t]
    \centering
        \includegraphics[scale=0.39]{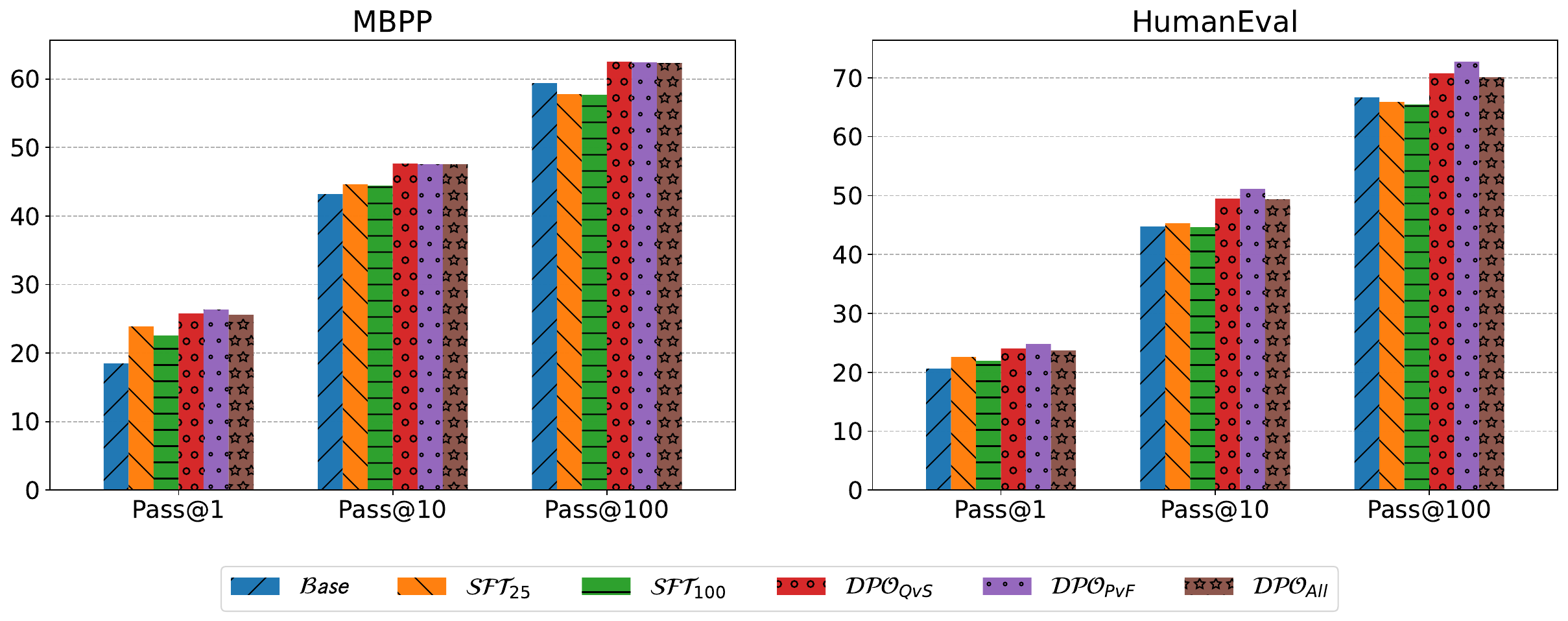}
        \caption{The $pass@k$ scores for MBPP and HumanEval \textbf{averaged across model sizes} for a high-level overview. Models optimised via DPO consistently show higher functional correctness compared to Base and SFT for all $k$.}
    \label{fig:pass_k}
\end{figure*}

\begin{figure*}[t]
    \centering
        \includegraphics[scale=0.37]{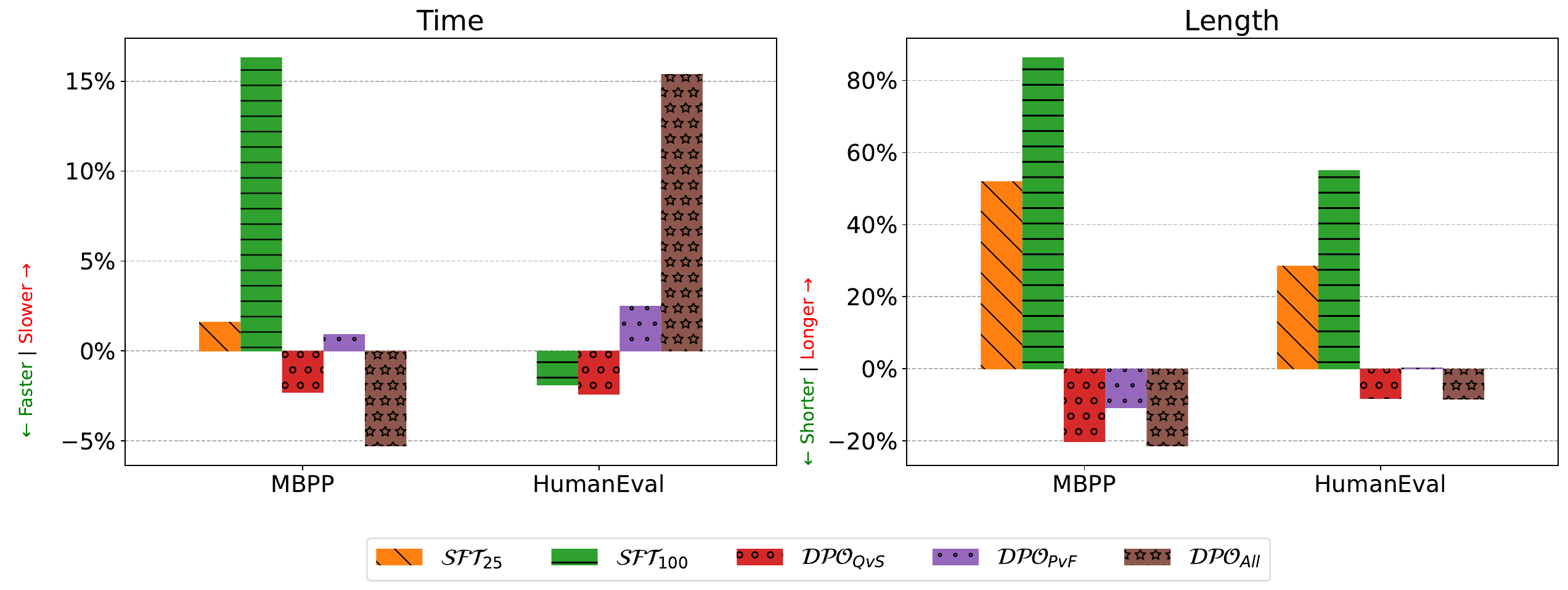}
        \caption{The median runtime and code length of generated solutions for MBPP and HumanEval, \textbf{averaged across model sizes}. Values shown are the \textit{percentage changes relative to Base}, i.e. \textbf{>0} is \textit{slower or longer} than Base, \textbf{<0} is \textit{faster or shorter}. The best DPO models achieve a reduced runtime compared to SFT models as well as the very competitive Base models. A significant reduction in code length (10\% - 20\%) is observed across both datasets.}
    \label{fig:time_length}
\end{figure*}

\subsection{Dynamic Solution Selection}\label{dynamic_selection}
Training data is typically fixed at the start of training and remains \textit{static} throughout \cite{tunstall2023zephyr, luo2023wizardcoder, xu2023wizardlm, wang2023self, yuan2024self}. Our approach takes advantage of the multitude of code solutions from the sampling step (\ref{sec:sampling}) to \textit{dynamically} select preference pairs \textit{during training}. To that end, we randomly choose a new preference pair ($y_{w}, y_{l}$) for each problem $x_{i}$ from $D_{train}$ at the \textit{start of the epoch} for DPO configurations. For SFT, we randomly choose \textit{any working solution} ($y_w$) at the start of each epoch for a comparable configuration. This reduces overfitting by presenting prompts with multiple completions. Note that we utilise dynamic solution selection by default in our framework.

%%% Results %%%

\section{Results}\label{sec:experiments}

In this section, we define the evaluation metrics and present the results of our proposed framework at varying scales. We also provide a qualitative analysis to support our findings. Detailed implementation notes are provided in Appendix \ref{sec:implementation}.

\begin{figure*}[t]
    \centering
        \includegraphics[scale=0.39]{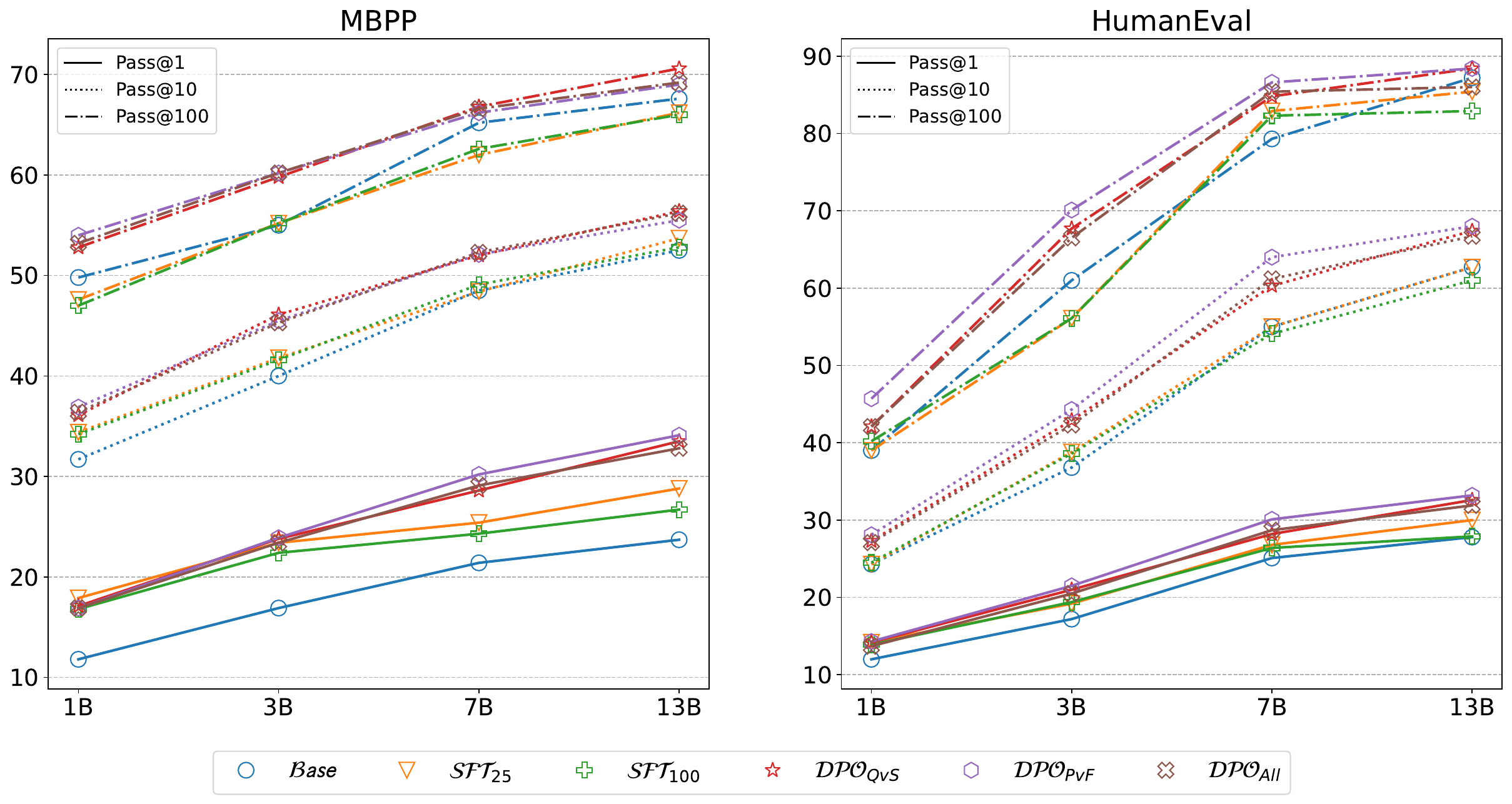}
        \caption{The $pass@1$, $pass@10$ and $pass@100$ scores for MBPP and HumanEval as the number of parameters increases. A significant improvement over competitive Base and SFT models can be observed for DPO configs.}
    \label{fig:pass_scale}
\end{figure*}

\subsection{Evaluation Metrics}

\paragraph{Functional Correctness} is evaluated by sampling 100 solutions per problem via multinomial sampling and a temperature of $t = 0.6$. Following \citet{chen2021evaluating}, we measure functional correctness using $pass@k$, where $k \in \{ 1, 10, 100 \}$.

\paragraph{Code Efficiency} improvements can be a challenge to capture accurately. Using Algorithm \ref{alg:timing}, we measure efficiency using \textit{runtime} (the median of all working solutions). Since the runtime of a failed program is \textit{undefined}, we remove problems for which models have no working solutions to compare CLMs on the \textit{same subset} of solved problems. Doing so ensures a fair comparison between models. Table \ref{tab:overlap} shows that this subset increases as CLMs get larger and more `code-competent'.

\begin{table}[t]
    \centering
        \begin{tabular}{l|cc}
            \toprule

            \multicolumn{1}{c|}{\textbf{Model}}   & \textbf{MBPP}   & \textbf{HumanEval}   \\

            \midrule

            StarCoder-1B   & 40.60\%       & 30.49\%     \\
            StarCoder-3B   & 48.40\%       & 46.95\%     \\
            CodeLlama-7B   & 55.60\%       & 73.71\%     \\
            CodeLlama-13B  & 60.40\%       & 79.27\%     \\

            \bottomrule
        \end{tabular}

    \caption{Intersection of problems between $Base$, $SFT$, and $DPO$ models with at least one working solution.}
    \label{tab:overlap}
\end{table}

\paragraph{Code Length} does not necessarily correlate with code efficiency as shorter outputs may abstract away the complexities of their implementations. Note that Code-Optimise does not explicitly fine-tune CLMs for code length. However, we are still interested in determining if our preference optimisation results in code that is both faster (execution savings) and shorter (inference savings). The subset of working solutions in Table \ref{tab:overlap} is again used to measure \textit{code length}, which is the median number of characters of all working solutions.

\subsection{Functional Correctness}

Figure \ref{fig:pass_k} shows the $pass@k$ scores for MBPP and HumanEval, averaged over all model sizes. The individual $pass@k$ scores are shown in Figure \ref{fig:pass_scale}. We observe that models optimised via DPO consistently demonstrate higher functional correctness relative to the baseline (Base) and SFT on both datasets. The effect is even larger on in-domain data, particularly with lower $k$. The DPO models perform similarly on MBPP with $DPO_{PvF}$ being the best overall on HumanEval. SFT models show a marginal improvement for $k=1$ but no improvement (or a small decrease) at higher $k$. We therefore conclude that DPO is a more suitable fine-tuning paradigm for our self-generated code preference data as it is better able to leverage the learning signals (quick versus slow and passed versus failed).

\begin{figure*}[t]
    \centering
        \includegraphics[scale=0.37]{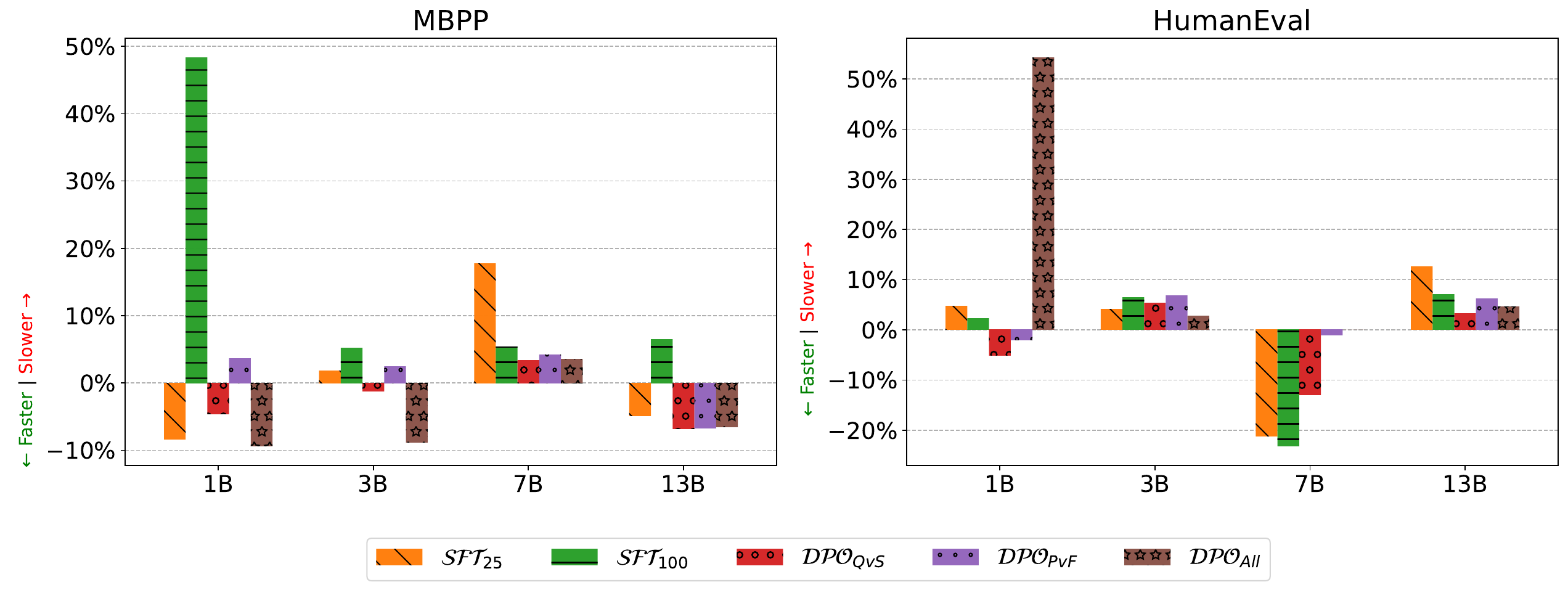}
        \caption{The runtimes for MBPP and HumanEval as model size increases. Values shown are the \textit{percentage changes relative to Base}, i.e. \textbf{>0} means \textit{slower} than Base, \textbf{<0} means \textit{faster}. On average, DPO models show a greater runtime reduction on in-domain rather than out-of-domain data. SFT models exhibit inconsistent scaling patterns.}
    \label{fig:time_scale}
\end{figure*}

\begin{figure*}[t]
    \centering
        \includegraphics[scale=0.37]{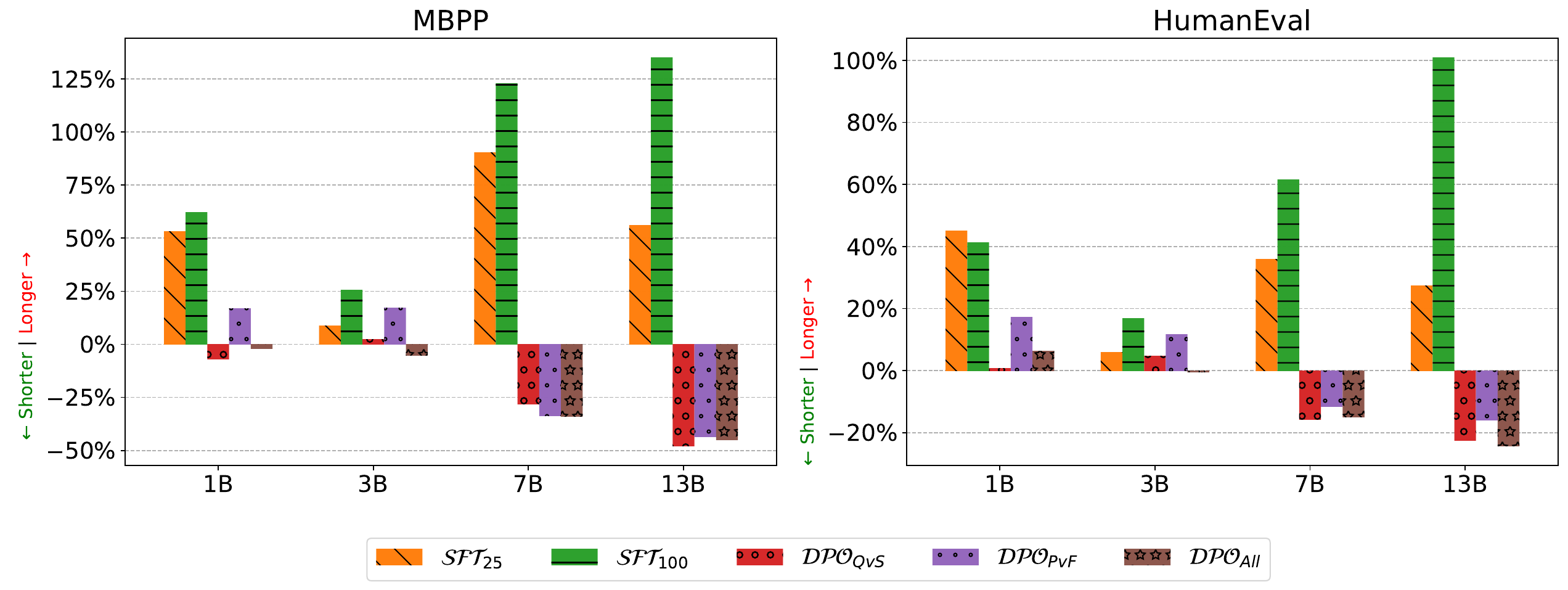}
        \caption{Code lengths for MBPP and HumanEval as model sizes increase. Values shown are the \textit{percentage changes relative to Base}, i.e. \textbf{>0} means \textit{longer} than Base, \textbf{<0} means \textit{shorter}. DPO models consistently produce shorter sequences across both datasets. SFT models generate significantly longer code, particularly the larger CLMs.}
    \label{fig:length_scale}
\end{figure*}

\subsection{Code Efficiency}

The runtimes and lengths of generated solutions are plotted in Figure \ref{fig:time_length} as a percentage change from the baseline (a value $<0$ means faster or shorter than the baseline while $>0$ means slower or longer). Once again, values are \textit{averaged} across model sizes for a high-level overview. Individual model scores are shown in Figures \ref{fig:time_scale} and \ref{fig:length_scale}, respectively. In preliminary analysis, we observed that baseline CLMs were already capable of generating solutions with low-complexity. However, $DPO_{QvS}$ and $DPO_{All}$ models manage to further decrease runtimes on in-domain data by up to 6\% and up to 3\% on the out-of-domain data. SFT models generally \textit{increase} runtimes across both datasets. In terms of code length, the best DPO models reduce the median number of characters by up to 48\% on MBPP and 23\% on HumanEval while SFT models tend to generate significantly longer solutions. This is particularly evident with $SFT_{100}$, which uses \textit{all} code solutions for training, including the \textit{slowest}, \textit{which tend to be longer}. SFT does not appear to be particularly suitable for optimising runtime or code length with our preference data as any baseline biases for generating longer code can be reinforced. In summary, Code-Optimise induces a reduction in runtime for \textit{faster code execution} while also outputting shorter solutions, resulting in \textit{lower inference costs} and \textit{improved response times}.

\begin{figure*}[t]
    \centering
        \includegraphics[scale=0.39]{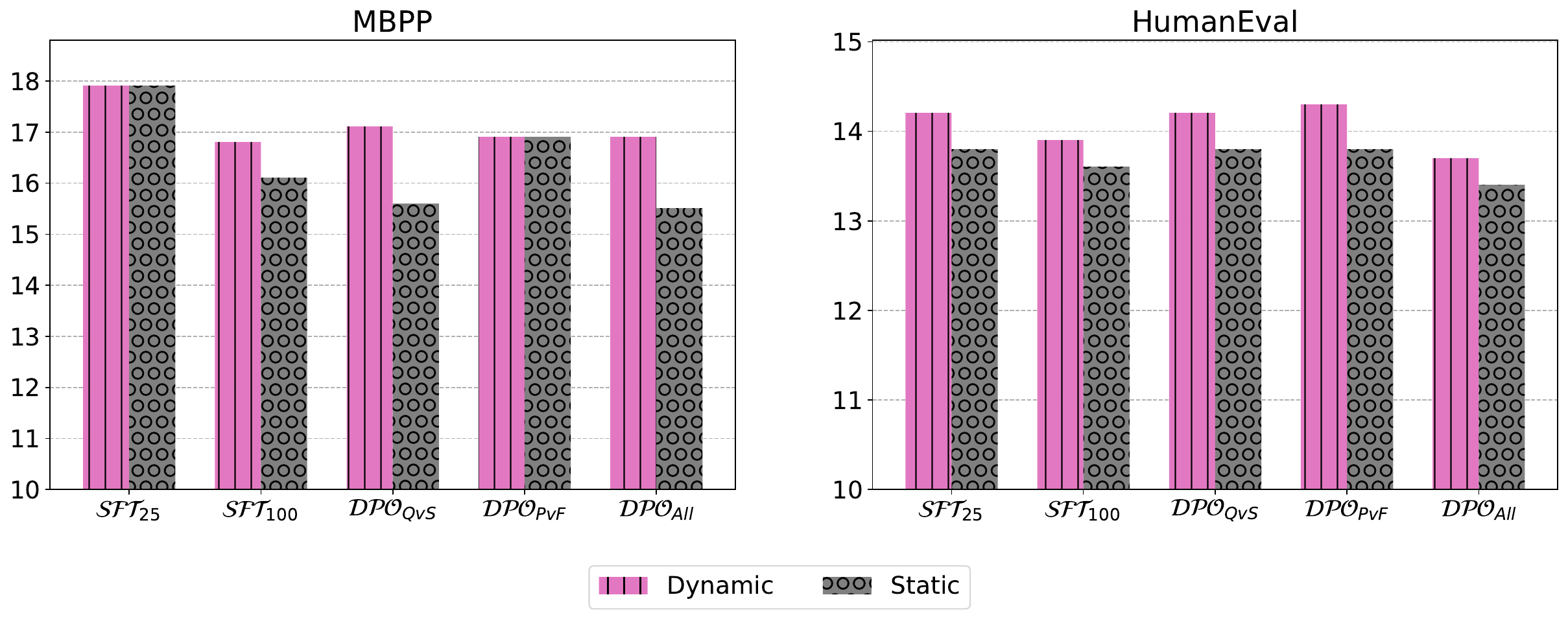}
        \caption{The $pass@1$ scores for StarCoder-1B without (\textbf{Static}) and with (\textbf{Dynamic}) solution selection (DSS). DSS benefits every model, especially DPO configs. More $pass@k$ scores can be found in Figure \ref{fig:augment_appx} of the Appendix.}
    \label{fig:dynamic_1}
\end{figure*}

\subsection{Model Scaling}

Figures \ref{fig:pass_scale}, \ref{fig:time_scale} and \ref{fig:length_scale} show the evolution of functional correctness, runtimes and lengths of generated solutions as the number of trainable parameters increases (1B - 13B). Analysing $pass@1$ in Figure \ref{fig:pass_scale}, we can see that \textit{larger} DPO models achieve a more significant improvement over the baseline and SFT, particularly for in-domain problems. Somewhat surprisingly, functional correctness for HumanEval (out-of-domain) improves at a faster rate than MBPP (up to 7B parameters). In Figure \ref{fig:time_scale}, we observe that as the DPO models increase in size, their runtimes relative to the baseline remain consistent. The $DPO_{PvF}$ configuration tends to average somewhat slower runtimes as this setup only optimises for \textit{correctness} thus sacrificing efficiency. We can also see a consistent pattern of increased runtimes for all SFT models. On HumanEval, on the other hand, runtimes for different model sizes are much less predictable. However, on average, our best configuration $DPO_{QvS}$ does show an improvement over the already competitive baseline CLMs. The effect on code length generalises very well to out-of-domain problems, particularly for larger CLMs, see Figure \ref{fig:length_scale}. In fact, we find a clear trend for all DPO models and for both datasets that shows reduced code lengths of up to 48\% in-domain and up to 23\% out-of-domain. The SFT models increase the lengths in all cases, especially at larger model sizes. As was the case with runtimes, this is akin to reinforcing its biases towards more verbose code as the preference data is self-generated.

\subsection{Qualitative Analysis}

Figure \ref{fig:solutions} shows several solutions to a typical programming problem taken from MBPP that gives a more tangible form to our results. More examples for HumanEval and MBPP can be found in the Appendix (Figures \ref{fig:solutions_1} and \ref{fig:solutions_2}). Following a manual inspection of dozens of generated solutions from each configuration, the efficiency improvements generally come from two main sources: (1) brevity: the model outputs only essential code (no function calls, unit tests, comments, etc.), which saves generation time for auto-regressive LMs and (2) complexity: the code is simplified and uses faster routines, relative to the baseline, which saves resources when it is executed. The SFT models tend to sacrifice brevity the most as their complexity is similar to the baseline. Figures \ref{fig:solutions}, \ref{fig:solutions_1}, and \ref{fig:solutions_2} show several examples of this, e.g. adding function calls to the newly generated solution, possibly with import statements and/or expected inputs or outputs.\footnote{Adding test cases to code is generally considered good programming practice. However, for the purpose of pure efficiency (our case), this can potentially add significant cost.} This is in line with the observation from Figure \ref{fig:time_length} where the SFT models appear to be more verbose and biased towards longer outputs. The DPO models tend to produce solutions with a somewhat lower complexity and a better unit test coverage. Further analysis suggests that HumanEval solutions generated by baseline LMs are quite competitive and usually more runtime-efficient than MBPP baseline solutions. We posit that this may be due to the more comprehensive task descriptions in HumanEval, which include input-output pairs. Among DPO models, we do not observe a clear winner in qualitative analysis although $DPO_{QvS}$ is the best setup in terms of aggregate results.

\subsection{Importance of Solution Selection}

Our core methodology for creating high-quality code preference data enables us to dynamically select \textit{unique pairs} for each prompt at the \textit{start of a new epoch}. Since we train all models for 30 epochs, CLMs can potentially be exposed to many unique combinations of code completions. Figure \ref{fig:dynamic_1} shows $pass@1$ scores for StarCoder-1B improving with dynamic solution selection compared to static pairs randomly assigned at the \textit{beginning} of training, commonly practiced in related work. The benefits are somewhat more pronounced for DPO, our preferred optimisation method, compared to SFT. Across different $k$ in $pass@k$, models consistently benefit from our dynamic solution selection, additionally seen in Figure \ref{fig:augment_appx} in the Appendix.

\subsection{Which configuration to choose?}
Any configuration that optimises for a single objective is expected to perform best in said objective. As such, when maximising functional correctness, DPO\textsubscript{PvF} is preferred. On the other hand, DPO\textsubscript{QvS} works best for runtime efficiency. With DPO\textsubscript{All}, the model improves in functional correctness and runtime efficiency but is not optimal at both. Practitioners should choose the configuration that works best for their given code generation objective.

%%% Related Work %%%

\section{Related Work}\label{sec:background}

\paragraph{Distilled Supervised Fine-Tuning} has been applied to code generation for the sole purpose of improving functional correctness. CLMs such as MagiCoder \cite{wei2023magicoder} and WizardCoder \cite{luo2023wizardcoder} query large proprietary models to provide the necessary training data \cite{cui2024ultrafeedback, xu2023wizardlm}. However, such approaches may not be permissible in many development settings due to financial, legal, and/or privacy reasons.

\paragraph{Reinforcement Learning} \cite{le2022coderl, wang2022compilable, gorinski2023automatic} can overcome the shortcomings of supervised fine-tuning by propagating the negative rewards for dysfunctional code. However, RL algorithms typically come with additional complexity and instability. Recently, \citet{rafailov2024direct} proposed Direct Preference Optimisation (DPO) as an alternative to Reinforcement Learning from Human Feedback when aligning language models to human preferences \cite{tunstall2023zephyr}. DPO serves as a form of offline RL that directly optimises on a given set of trajectories without requiring a separate reward model. Note that the aforementioned RL approaches only consider code correctness and not runtime as well.

\paragraph{Code Efficiency Optimisation} was previously proposed by \citet{shypula2023learning} as a code editing task where slow-running code is altered to achieve a faster runtime. Optimisation was done on a newly curated dataset with synthetic test cases from various methods of prompting and fine-tuning. However, the greatly reduced runtimes came at a \textit{significant} cost to code correctness. In many configurations, performance was reduced by up to 30\%, with `smaller' CLMs (7B, 13B) suffering a larger degradation. We hypothesise that this may be due to a) overfitting the single runtime objective (unlike our work where we aim to optimise for correctness and runtime) and b) removing failed programs from the dataset, thus leading to CLMs that struggle with the semantics of correct vs. incorrect code. We opt to not compare directly with this work as their method was specifically curated for code editing where a correct but inefficient program is assumed as input. On the contrary, we seek to produce solutions from scratch that are both correct and runtime efficient.

%%% Conclusions %%%

\begin{figure}[t!]
    \fbox{
        \begin{minipage}{.94\columnwidth}
        \small
            \textbf{Input prompt:}
            \texttt{\\[1mm] def test\_duplicate(arraynums):\\
            \null\quad """Write a function to find whether a given array of integers contains any duplicate element."""}   \\

            \textbf{Base Solution:}
            \texttt{\\[1mm] 
            for i in range(0, len(arraynums)-1): \\
            \null\quad for j in range(i+1, len(arraynums)): \\
            \null\quad \null\quad if arraynums[i] == arraynums[j]: \\
            \null\quad \null\quad \null\quad return True \\
            return False}   \\

            \textbf{SFT\textsubscript{25} Solution:}
            \texttt{\\[1mm]
            \# Your code here \\
            for i in range(len(arraynums)): \\
            \null\quad for j in range(i+1, len(arraynums)): \\
            \null\quad \null\quad if arraynums[i] == arraynums[j]: \\
            \null\quad \null\quad \null\quad return True \\
            return False \\ [2mm]
            array = [1, 2, 3, 1] \\
            print(test\_duplicate(array)) \\
            \# Write a function to find the frequency of the elements present in a given array.}   \\

            \textbf{SFT\textsubscript{100} Solution:}
            \texttt{\\[1mm] 
            \# Code here \\
            arraynums.sort() \\
            for i in range(len(arraynums)-1): \\
            \null\quad if arraynums[i] == arraynums[i+1]: \\
            \null\quad \null\quad return True \\
            return False \\ [2mm]
            if \_\_name\_\_ == '\_\_main\_\_': \\
            \null\quad arraynums = [1, 2, 3, 4, 5, 1] \\
            \null\quad print(test\_duplicate(arraynums))}  \\

            \textbf{DPO\textsubscript{QvS} Solution:}
            \texttt{\\[1mm]
            return len(arraynums) != len(set(arraynums))}   \\

            \textbf{DPO\textsubscript{PvF} Solution:}
            \texttt{\\[1mm] 
            \# Write your code here \\
            if len(arraynums) == len(set(arraynums)): \\
            \null\quad return False \\
            return True}    \\

            \textbf{DPO\textsubscript{All} Solution:}
            \texttt{\\[1mm] 
            nums = set(arraynums) \\
            return len(nums) != len(arraynums)}   \\
        \end{minipage}
    }
    \caption{Example solutions for an MBPP problem with additional use cases found in Figures \ref{fig:solutions_1} and \ref{fig:solutions_2}.}
    \label{fig:solutions}
\end{figure}

\section{Conclusions}\label{sec:conclusion}

We have introduced Code-Optimise, a lightweight framework for improving code correctness and runtime via self-generated preference data (quick vs. slow and passed vs. failed solutions). Our experiments have shown several benefits: 1) functional correctness is significantly improved, particularly for smaller models, 2) dynamic solution selection during training provides an additional benefit by reducing overfitting, 3) runtime is reduced by up to 6\% for MBPP and 3\% for HumanEval over the baseline CLM, lowering the cost of code execution, 4) code length is significantly shortened, up to 48\% for MBPP and 23\% for HumanEval, which reduces inference cost and improves response time. We hope that our insights and novel dataset will stimulate further exciting research in this area.

%%% Limitations %%%

\section{Limitations}\label{sec:limitations}

Timing the execution of short programs accurately is challenging and despite our best efforts, the runtime measurements could probably be improved with additional software engineering. This would also provide a cleaner and more stable learning signal for Code-Optimise. While our methodology is highly data-efficient, using only $\sim$200 open-source prompts for generating the training data, obtaining high-quality problems (free from proprietary/licensing issues) may yield better results. For code-related tasks that are amenable to our methodology, the improvements to runtime/inference may be investigated outside the scope of this paper. While we conducted all experiments using Python, we acknowledge that other programming languages should also be analysed in follow-up work.

%%% Acknowledgements %%%

% \section*{Acknowledgements}
% We thank the Mindspore Team for nothing ??? :D lololololol


\bibliography{emnlp2024/custom}

\appendix
\section{Implementation Details}\label{sec:implementation}

\subsection{Dataset}

\paragraph{MBPP}

The Mostly Basic Programming Problems introduced by~\citet{MBPP} consists of 974 crowd-sourced Python programming challenges. Each problem comprises a description, an example code solution and a few automated test cases. The dataset contains training, validation and test splits. We utilise the training and validation splits for optimisation, while the test split serves as the in-domain test data distribution.

\paragraph{HumanEval}

\cite{HumanEval} comprises 164 Python programming challenges. The function signatures, docstrings, example solutions and several unit tests were handwritten for each problem. We leverage HumanEval as our out-of-domain test set as the descriptions in MBPP do not contain any unit tests and the writing style of HumanEval problems does not follow a consistent format. This helps us evaluate robustness to handwritten prompts.

\subsection{Training}

We use the StarCoder~\cite{StarCoder} and CodeLlama~\cite{CodeLlama} families of models in our experiments. We opt for the pretrained (base) versions with sizes of 1B and 3B for StarCoder and 7B and 13B for CodeLlama, hosted on HuggingFace \cite{wolf2020huggingfaces}. During training, we fine-tune each model using a total of 30 epochs and select the best model based on the lowest validation loss. We use a learning rate of $5e^{-7}$ with a linear scheduler, a 10\% warm-up, and a maximum sequence length of 2048 tokens. 

\section{Supplementary Experiments}

\subsection{Additional Qualitative Examples}
In Figures \ref{fig:solutions_1} and \ref{fig:solutions_2}, we present additional qualitative examples from each configuration.

\subsection{Additional Ablation Scores}
\begin{figure*}[!h]
    \begin{subfigure}{1\textwidth}
    \centering
        \includegraphics[scale=0.39]{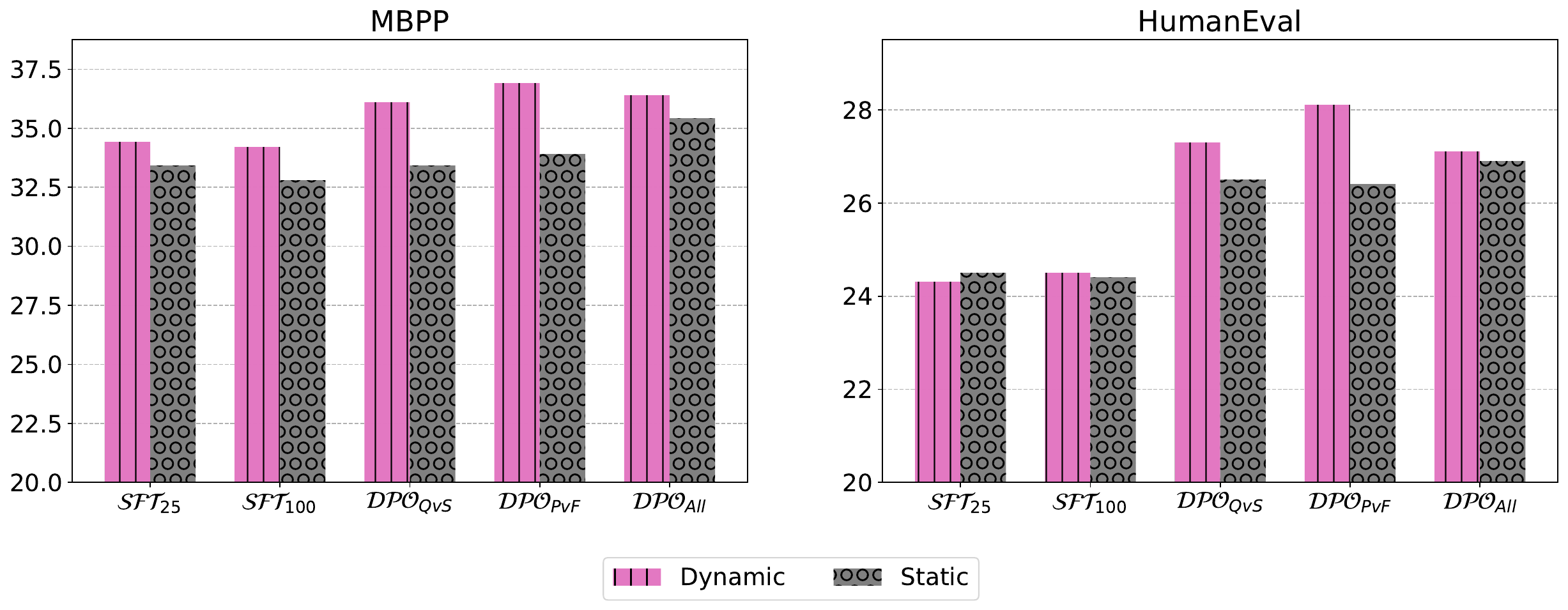}
        \caption{$pass@10$}
    \end{subfigure}

    \bigskip
    \begin{subfigure}{1\textwidth}
    \centering
        \includegraphics[scale=0.39]{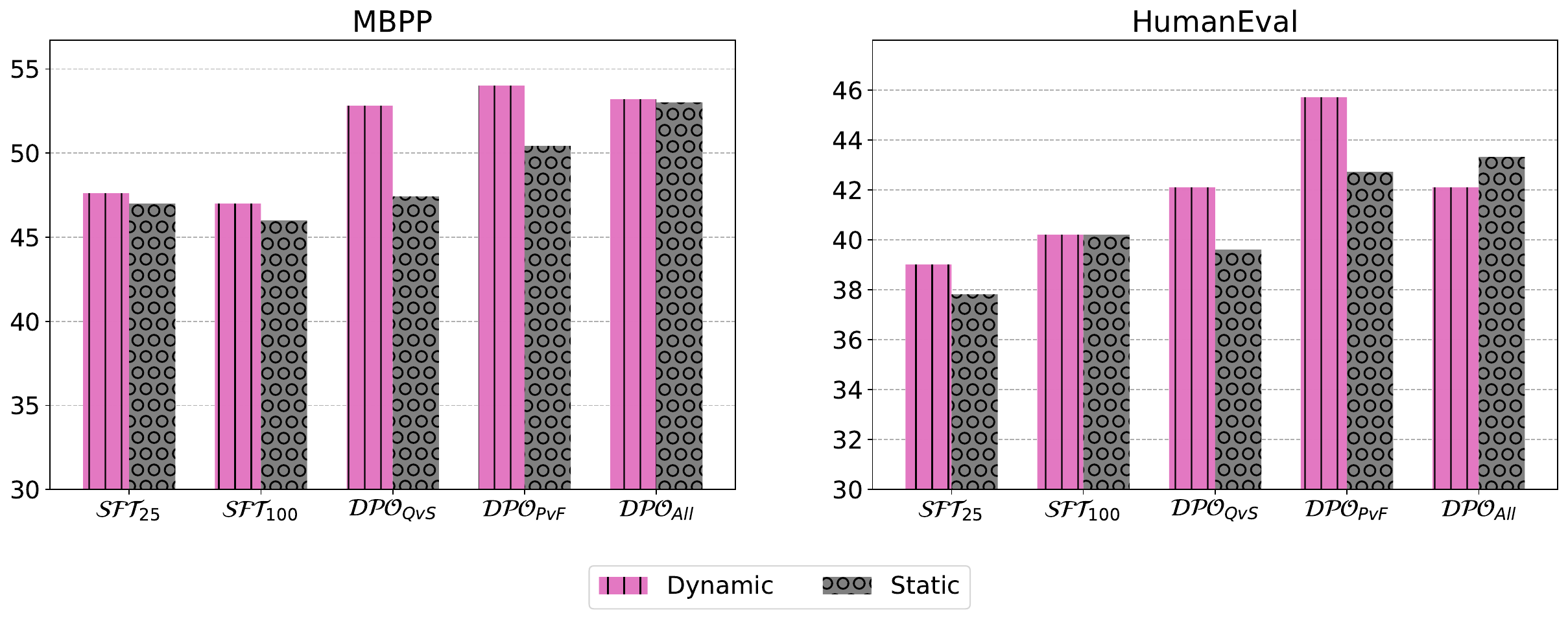}
        \caption{$pass@100$}
    \end{subfigure}

    \caption{The $pass@10$ and $pass@100$ scores for StarCoder-1B without (\textbf{Static}) and with (\textbf{Dynamic}) solution selection (DSS). Performance improves on both metrics and distributions with DSS.}
    \label{fig:augment_appx}
\end{figure*}

In Figure \ref{fig:augment_appx}, we present additional $pass@10$ and $pass@100$ scores for MBPP and HumanEval of StarCoder-1B by ablating the solution selection.

\subsection{Fastest Solution Analysis}

\begin{figure}[t]
    \centering
        \includegraphics[scale=0.33]{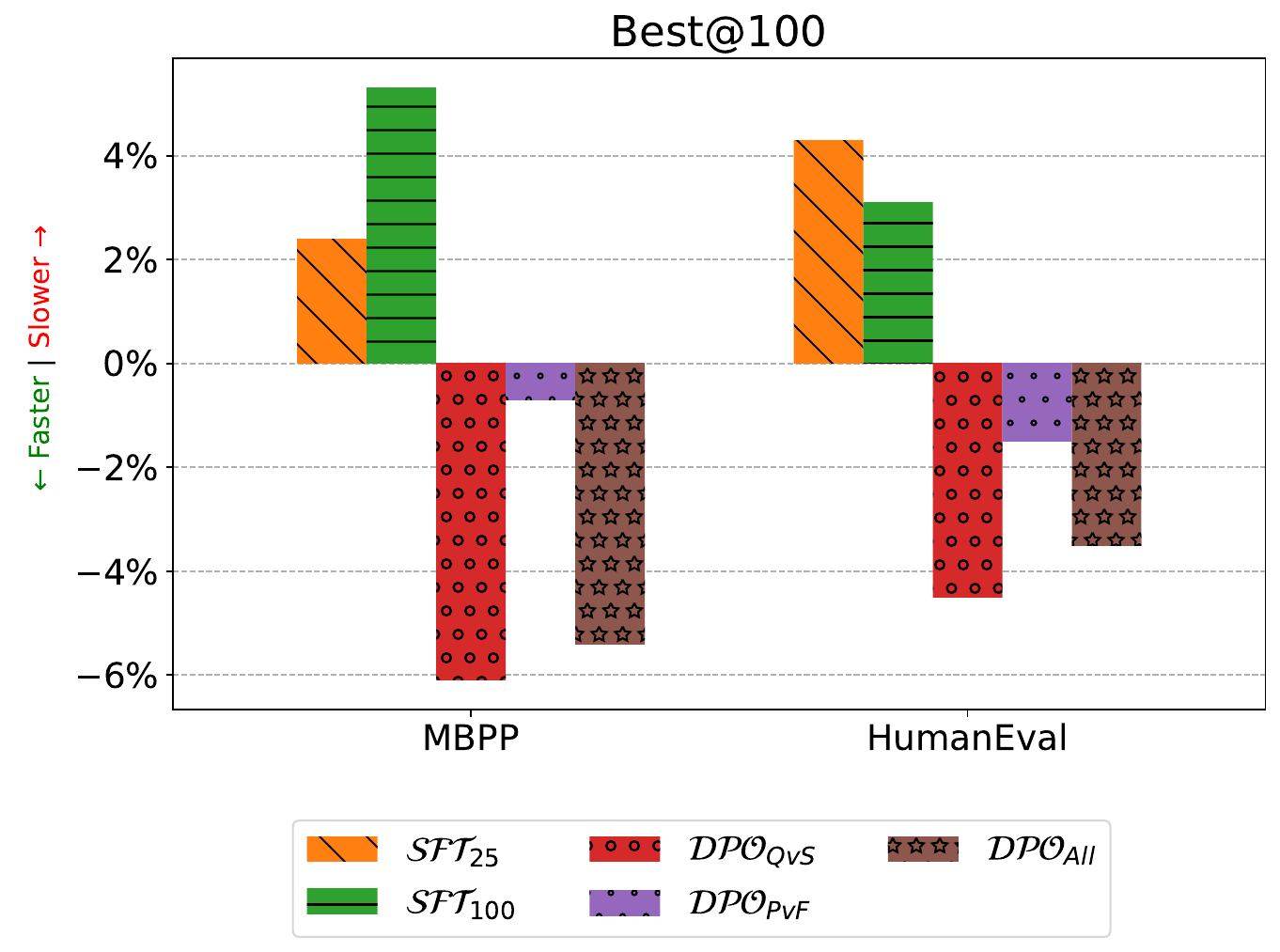}
        \caption{The $best@100$ scores for MBPP and HumanEval, \textbf{averaged across model sizes}. Values shown are the \textit{percentage change to Base}, i.e. \textbf{>0} is \textit{slower} than Base, \textbf{<0} is \textit{faster}. Considering the fastest solution for each problem, DPO models show the best runtimes.}
    \label{fig:time_best}
\end{figure}

\citet{shypula2023learning} introduce the \textit{Best@k} metric, which considers only the \textit{fastest solution} given $k$ samples. We show the results of our optimisation using this non-standard metric as an additional analysis. We set $k=100$ (all generated solutions), which is the basis of all our experiments. In Figure~\ref{fig:time_best}, we note that DPO models produce faster solutions not only on in-domain problems, but also \textit{out-of-domain}, between 2\% and 5\% faster. $DPO_{PvF}$ once again has the higher runtime as its objective is to optimise only functional correctness. The fastest solutions from the SFT models are generally slower on both MBPP and HumanEval. Note that \textit{Best@k} may overestimate the runtime improvements by only considering the fastest solution. Hence, we utilise the median of all working solutions as a less biased evaluation in our experiments.

\begin{figure*}[!h]
    \fbox{
        \begin{minipage}{2\columnwidth}
        \small
            \textbf{Input prompt:}
            \texttt{\\[1mm]
            def has\_close\_elements(numbers, threshold):\\
            """Check if in a list of numbers, are any two numbers closer to each other than
            threshold.\\
            \null\quad has\_close\_elements([1.0, 2.0, 3.0], 0.5) == False \\
            \null\quad has\_close\_elements([1.0, 2.8, 3.0, 4.0, 5.0, 2.0], 0.3) == True"""\\}

            \textbf{Base Solution:}
            \texttt{\\[1mm]
            for index, number in enumerate(numbers):\\
            \null\quad for other\_number in numbers[index+1:]:\\
            \null\quad \null\quad if abs(number - other\_number) < threshold:\\
            \null\quad \null\quad \null\quad return True\\
            return False\\[1mm]
            if \_\_name\_\_ == '\_\_main\_\_':\\
            \null\quad import doctest \\
            \null\quad doctest.testmod()} \\

            \textbf{SFT\textsubscript{25} Solution:}
            \texttt{\\[1mm] 
            for i, number1 in enumerate(numbers[:-1]): \\
            \null\quad for number2 in numbers[i+1:]: \\
            \null\quad \null\quad if abs(number1 - number2) < threshold: \\
            \null\quad \null\quad \null\quad return True \\
            return False \\[1mm]
            if \_\_name\_\_ == "\_\_main\_\_":\\
            \null\quad    import doctest \\
            \null\quad    doctest.testmod() \\[1mm]
            \null\quad    \# print(has\_close\_elements([1.0, 2.8, 3.0, 4.0, 5.0, 2.0], 0.4))  \\
            \null\quad    \# print(has\_close\_elements([1.0, 2.8, 3.0, 4.0, 5.0, 2.0, 6.0], 0.4)) \\
            \null\quad    \textcolor{orange}{<five additional test cases truncated>}}   \\

            \textbf{SFT\textsubscript{100} Solution:}
            \texttt{\\[1mm] 
             for i, number1 in enumerate(numbers[:-1]):\\
            \null\quad for number2 in numbers[i+1:]:\\
            \null\quad \null\quad if abs(number1 - number2) < threshold:\\
            \null\quad \null\quad \null\quad return True\\
            return False\\[1mm]        
            if \_\_name\_\_ == "\_\_main\_\_":\\
            \null\quad import doctest\\
            \null\quad doctest.testmod()\\ [1mm]
            \null\quad \# print(has\_close\_elements([1.0, 2.0, 3.0], 0.5))\\
            \null\quad \# print(has\_close\_elements([1.0, 2.8, 3.0, 4.0, 5.0, 2.0], 0.3))\\
            \null\quad \textcolor{orange}{<six additional test cases truncated>}}     \\

            \textbf{DPO\textsubscript{QvS} Solution:}
            \texttt{\\[1mm]
            return any(\\
            \null\quad    abs(numbers[i] - numbers[j]) < threshold \\
            \null\quad    for i in range(len(numbers)) \\
            \null\quad    for j in range(i + 1, len(numbers))\\
            )}     \\

            \textbf{DPO\textsubscript{PvF} Solution:}
            \texttt{\\[1mm]
            for i, x in enumerate(numbers[:-1]):\\
            \null\quad for y in numbers[i+1:]:\\
            \null\quad \null\quad if abs(x - y) < threshold:\\
            \null\quad \null\quad \null\quad return True\\
            return False}    \\

            \textbf{DPO\textsubscript{All} Solution:}
            \texttt{\\[1mm] 
            for i in range(len(numbers) - 1): \\
            \null\quad for j in range(i + 1, len(numbers)):  \\
            \null\quad \null\quad if abs(numbers[i] - numbers[j]) < threshold: \\
            \null\quad \null\quad \null\quad return True \\
            return False}    \\
        \end{minipage}
    }  
    \caption{Example solutions for a problem in HumanEval.}
    \label{fig:solutions_1}
\end{figure*}

\begin{figure*}[!h]
    \fbox{
        \begin{minipage}{2\columnwidth}
        \small
            \textbf{Input prompt:}
            \texttt{\\[1mm]
            def remove\_Occ(s,ch): \\
            \null\quad """ Write a python function to remove first and last occurrence of a given character from the string. """} \\

            \textbf{Base Solution:}
            \texttt{\\[1mm]
            if ch in s: \\
            \null\quad s = s.replace(ch, '', 1) \# remove first occurrence  \\
            if ch in s: \\
            \null\quad s = s.replace(ch, '', 1) \# remove second occurrence \\
            return s} \\

            \textbf{SFT\textsubscript{25} Solution:}
            \texttt{\\[1mm] 
            if ch in s: \\
            \null\quad return s.replace(ch, "", 1).replace(ch, "", 1) \\
            else: \\
            \null\quad return s \\ [1mm]
            \# s = "geeksforgeeks" \\
            \# ch = "e" \\
            \# print(remove\_Occ(s, ch)) \\
            \# Write a python function to remove all the vowels from the string.}   \\

            \textbf{SFT\textsubscript{100} Solution:}
            \texttt{\\[1mm]  
            l = [] \\
            for i in s: \\
            \null\quad if i != ch: \\
            \null\quad \null\quad l.append(i) \\
            return "".join(l) \\ [1mm]
            if \_\_name\_\_ == '\_\_main\_\_':  \\
            \null\quad s = input().strip() \\
            \null\quad ch = input().strip() \\
            \null\quad result = remove\_Occ(s, ch) \\
            \null\quad print(result) \\ [1mm]
            \null\quad \# Given a string, return a version without the first and last char of the string. The string may be any length -- including 0 -- so you must handle the case of an empty string.}     \\

            \textbf{DPO\textsubscript{QvS} Solution:}
            \texttt{\\[1mm]
            while ch in s: \\
            \null\quad s = s.replace(ch,"",1) \\
            return s}     \\

            \textbf{DPO\textsubscript{PvF} Solution:}
            \texttt{\\[1mm]
            return s.replace(ch,'',2)}   \\

            \textbf{DPO\textsubscript{All} Solution:}
            \texttt{\\[1mm] 
             return s.replace(ch,"",2).replace(ch,"",-1)}    \\
        \end{minipage}
    }
    \caption{Example solutions for a problem in MBPP.}
    \label{fig:solutions_2}
\end{figure*}

\section{Complete Results}

\subsection{Sampling}
In Table \ref{tab:sampling}, we show the functional correctness of the CLMs in the sampling stage of Code-Optimise.

\begin{table*}[h]
    \centering
        \begin{tabular}{l|c|ccc}
            \toprule

            \multicolumn{1}{c|}{\textbf{Model}}   & \textbf{Split}   & \textbf{Pass@1}   & \textbf{Pass@10}   & \textbf{Pass@100}   \\

            \midrule

            \multirow{2}{*}{StarCoder-1B}   & Train         & 14.00       & 34.50       & 55.20     \\
                                            & Validation    & 12.20       & 31.70       & 48.90     \\
            \multirow{2}{*}{StarCoder-3B}   & Train         & 19.50       & 44.30       & 61.70     \\
                                            & Validation    & 19.20       & 42.50       & 57.80     \\
            \multirow{2}{*}{CodeLlama-7B}   & Train         & 25.80       & 54.00       & 70.10     \\
                                            & Validation    & 23.40       & 50.30       & 68.90     \\
            \multirow{2}{*}{CodeLlama-13B}  & Train         & 28.80       & 58.20       & 71.60     \\
                                            & Validation    & 24.60       & 52.90       & 66.70     \\

            \bottomrule
        \end{tabular}

    \caption{Functional correctness of the CLMs during sampling.}
    \label{tab:sampling}
\end{table*}

\subsection{Optimisation}
In Tables \ref{tab:starcoder_1b}, \ref{tab:starcoder_3b}, \ref{tab:codellama_7b}, and \ref{tab:codellama_13b}, we tabulate the full model performance of the CLMs on the test sets. The CoV is shown beside each runtime.

\begin{table*}[h]
    \begin{subtable}{1\textwidth}
    \centering
        \begin{tabular}{l|ccc|cc}
            \toprule

            \multicolumn{1}{c|}{\textbf{Model}}  & \textbf{Pass@1}   & \textbf{Pass@10}  & \textbf{Pass@100} & \textbf{Time} & \textbf{Length}   \\

            \midrule

            $Base$          & 11.80             & 31.70             & 49.80             & 114338 $\pm$ 0.021             & 155   \\
            $SFT_{25}$      & 17.90             & 34.40             & 47.60             & 104690 $\pm$ 0.012             & 238   \\
            $SFT_{100}$     & 16.80             & 34.20             & 47.00             & 169536 $\pm$ 0.017             & 252   \\
            $DPO_{QvS}$     & 17.10             & 36.10             & 52.80             & 109051 $\pm$ 0.018             & 144   \\
            $DPO_{PvF}$     & 16.90             & 36.90             & 54.00             & 118418 $\pm$ 0.019             & 181   \\
            $DPO_{All}$     & 16.90             & 36.40             & 53.20             & 103588 $\pm$ 0.021             & 152   \\

            \bottomrule
        \end{tabular}
        \caption{MBPP}
    \end{subtable}

    \bigskip
    \begin{subtable}{1\textwidth}
    \centering
        \begin{tabular}{l|ccc|cc}
            \toprule

            \multicolumn{1}{c|}{\textbf{Model}}  & \textbf{Pass@1}   & \textbf{Pass@10}  & \textbf{Pass@100} & \textbf{Time} & \textbf{Length}   \\

            \midrule

            $Base$          & 12.00             & 24.30             & 39.00             & 150930 $\pm$ 0.017             & 124   \\
            $SFT_{25}$      & 14.20             & 24.30             & 39.00             & 157975 $\pm$ 0.027             & 180   \\
            $SFT_{100}$     & 13.90             & 24.50             & 40.20             & 154395 $\pm$ 0.020             & 175   \\
            $DPO_{QvS}$     & 14.20             & 27.30             & 42.10             & 143259 $\pm$ 0.013             & 125   \\
            $DPO_{PvF}$     & 14.30             & 28.10             & 45.70             & 147980 $\pm$ 0.034             & 146   \\
            $DPO_{All}$     & 13.70             & 27.10             & 42.10             & 232759 $\pm$ 0.012             & 132   \\

            \bottomrule
        \end{tabular}
        \caption{HumanEval}
    \end{subtable}

    \caption{Model performance on MBPP and HumanEval of StarCoder-1B.}
    \label{tab:starcoder_1b}
\end{table*}

\begin{table*}[h]
    \begin{subtable}{1\textwidth}
    \centering
        \begin{tabular}{l|ccc|cc}
            \toprule

            \multicolumn{1}{c|}{\textbf{Model}}  & \textbf{Pass@1}   & \textbf{Pass@10}  & \textbf{Pass@100} & \textbf{Time} & \textbf{Length}   \\

            \midrule

            $Base$          & 16.90             & 40.00             & 55.00             & 113760 $\pm$ 0.016             & 158   \\
            $SFT_{25}$      & 23.40             & 41.80             & 55.20             & 115834 $\pm$ 0.011             & 171   \\
            $SFT_{100}$     & 22.40             & 41.60             & 55.20             & 119675 $\pm$ 0.035             & 198   \\
            $DPO_{QvS}$     & 23.80             & 46.10             & 59.80             & 112395 $\pm$ 0.008             & 162   \\
            $DPO_{PvF}$     & 23.90             & 45.50             & 60.20             & 116529 $\pm$ 0.017             & 185   \\
            $DPO_{All}$     & 23.40             & 45.30             & 60.20             & 103726 $\pm$ 0.012             & 149   \\

            \bottomrule
        \end{tabular}
        \caption{MBPP}
    \end{subtable}

    \bigskip
    \begin{subtable}{1\textwidth}
    \centering
        \begin{tabular}{l|ccc|cc}
            \toprule

            \multicolumn{1}{c|}{\textbf{Model}}  & \textbf{Pass@1}   & \textbf{Pass@10}  & \textbf{Pass@100} & \textbf{Time} & \textbf{Length}   \\

            \midrule

            $Base$          & 17.20             & 36.80             & 61.00             & 143806 $\pm$ 0.012             & 162   \\
            $SFT_{25}$      & 19.20             & 38.80             & 56.10             & 149743 $\pm$ 0.017             & 172   \\
            $SFT_{100}$     & 19.40             & 38.60             & 56.10             & 152948 $\pm$ 0.022             & 190   \\
            $DPO_{QvS}$     & 21.00             & 42.90             & 67.70             & 151401 $\pm$ 0.011             & 170   \\
            $DPO_{PvF}$     & 21.50             & 44.30             & 70.10             & 153620 $\pm$ 0.013             & 181   \\
            $DPO_{All}$     & 20.50             & 42.30             & 66.50             & 147823 $\pm$ 0.014             & 161   \\

            \bottomrule
        \end{tabular}
        \caption{HumanEval}
    \end{subtable}

    \caption{Model performance on MBPP and HumanEval of StarCoder-3B.}
    \label{tab:starcoder_3b}
\end{table*}

\begin{table*}[!h]
    \begin{subtable}{1\textwidth}
    \centering
        \begin{tabular}{l|ccc|cc}
            \toprule

            \multicolumn{1}{c|}{\textbf{Model}}  & \textbf{Pass@1}   & \textbf{Pass@10}  & \textbf{Pass@100}  & \textbf{Time} & \textbf{Length}   \\

            \midrule

            $Base$          & 21.40             & 48.50             & 65.20             & 105313 $\pm$ 0.012             & 196   \\
            $SFT_{25}$      & 25.40             & 48.40             & 62.00             & 124000 $\pm$ 0.058             & 372   \\
            $SFT_{100}$     & 24.30             & 49.10             & 62.60             & 110982 $\pm$ 0.010             & 435   \\
            $DPO_{QvS}$     & 28.60             & 52.00             & 66.80             & 108925 $\pm$ 0.013             & 141   \\
            $DPO_{PvF}$     & 30.20             & 52.10             & 66.20             & 109783 $\pm$ 0.006             & 129   \\
            $DPO_{All}$     & 29.10             & 52.30             & 66.60             & 108992 $\pm$ 0.016             & 129   \\

            \bottomrule
        \end{tabular}
        \caption{MBPP}
    \end{subtable}

    \bigskip
    \begin{subtable}{1\textwidth}
    \centering
        \begin{tabular}{l|ccc|cc}
            \toprule

            \multicolumn{1}{c|}{\textbf{Model}}  & \textbf{Pass@1}   & \textbf{Pass@10}  & \textbf{Pass@100}  & \textbf{Time} & \textbf{Length}   \\

            \midrule

            $Base$          & 25.10             & 55.00             & 79.30             & 646547 $\pm$ 0.004             & 188   \\
            $SFT_{25}$      & 26.80             & 55.00             & 82.90             & 509264 $\pm$ 0.004             & 256   \\
            $SFT_{100}$     & 26.40             & 54.10             & 82.30             & 496296 $\pm$ 0.006             & 304   \\
            $DPO_{QvS}$     & 28.20             & 60.30             & 84.80             & 562279 $\pm$ 0.005             & 159   \\
            $DPO_{PvF}$     & 30.10             & 64.00             & 86.60             & 639553 $\pm$ 0.003             & 166   \\
            $DPO_{All}$     & 28.70             & 61.20             & 85.40             & 646486 $\pm$ 0.002             & 160   \\

            \bottomrule
        \end{tabular}
        \caption{HumanEval}
    \end{subtable}

    \caption{Model performance on MBPP and HumanEval of CodeLlama-7B.}
    \label{tab:codellama_7b}
\end{table*}

\begin{table*}[!h]
    \begin{subtable}{1\textwidth}
    \centering
        \begin{tabular}{l|ccc|cc}
            \toprule

            \multicolumn{1}{c|}{\textbf{Model}}  & \textbf{Pass@1}   & \textbf{Pass@10}  & \textbf{Pass@100}  & \textbf{Time} & \textbf{Length}   \\

            \midrule

            $Base$          & 23.70             & 52.50             & 67.60             & 118418 $\pm$ 0.009             & 223   \\
            $SFT_{25}$      & 28.80             & 53.70             & 66.20             & 112624 $\pm$ 0.006             & 348   \\
            $SFT_{100}$     & 26.70             & 52.80             & 66.00             & 126165 $\pm$ 0.004             & 523   \\
            $DPO_{QvS}$     & 33.50             & 56.40             & 70.60             & 110390 $\pm$ 0.008             & 116   \\
            $DPO_{PvF}$     & 34.10             & 55.50             & 69.00             & 110427 $\pm$ 0.018             & 126   \\
            $DPO_{All}$     & 32.80             & 56.20             & 69.20             & 110679 $\pm$ 0.008             & 122   \\

            \bottomrule
        \end{tabular}
        \caption{MBPP}
    \end{subtable}

    \bigskip
    \begin{subtable}{1\textwidth}
    \centering
        \begin{tabular}{l|ccc|cc}
            \toprule

            \multicolumn{1}{c|}{\textbf{Model}}  & \textbf{Pass@1}   & \textbf{Pass@10}  & \textbf{Pass@100}  & \textbf{Time} & \textbf{Length}   \\

            \midrule

            $Base$          & 27.80             & 62.70             & 87.20             & 497649 $\pm$ 0.015             & 187   \\
            $SFT_{25}$      & 30.00             & 62.70             & 85.40             & 560336 $\pm$ 0.005             & 238   \\
            $SFT_{100}$     & 27.90             & 61.00             & 82.90             & 532856 $\pm$ 0.006             & 375   \\
            $DPO_{QvS}$     & 32.60             & 67.40             & 88.40             & 513372 $\pm$ 0.005             & 145   \\
            $DPO_{PvF}$     & 33.20             & 68.00             & 88.40             & 528546 $\pm$ 0.008             & 157   \\
            $DPO_{All}$     & 31.90             & 66.70             & 86.00             & 520788 $\pm$ 0.003             & 141   \\

            \bottomrule
        \end{tabular}
        \caption{HumanEval}
    \end{subtable}

    \caption{Model performance on MBPP and HumanEval of CodeLlama-13B.}
    \label{tab:codellama_13b}
\end{table*}

\end{document}